\documentclass[11pt,a4paper]{article}
\usepackage[hyperref]{naaclhlt2019}
\usepackage{times}
\usepackage{latexsym}
\usepackage[colorinlistoftodos, textsize=tiny]{todonotes}
\usepackage{url}
\usepackage{multirow}
\usepackage{float}
\usepackage{mathtools}
\usepackage{amsfonts}
\usepackage{amsmath}
\usepackage{booktabs}
\usepackage{etoolbox}
\makeatletter
\patchcmd\@combinedblfloats{\box\@outputbox}{\unvbox\@outputbox}{}{}%\errmessage{\noexpand patch failed}} %% Throw error fallback commented
\makeatother
\DeclareMathOperator*{\argmax}{arg\,max}
\DeclareMathOperator*{\mean}{mean}
\DeclareMathOperator*{\std}{std}

\aclfinalcopy % Uncomment this line for the final submission
\def\aclpaperid{729} %  Enter the acl Paper ID here

\title{Cooperative Learning of Disjoint Syntax and Semantics}

\author{Serhii Havrylov \Thanks{Work done while the author was an intern at Facebook AI Research.}\\
ILCC, University of Edinburgh / Edinburgh, UK \\
{\tt s.havrylov@ed.ac.uk}\\\AND
Germ\'an Kruszewski \& Armand Joulin \\
Facebook AI Research \\
{\tt\{germank,ajoulin\}@fb.com} \\
}

\date{}

\begin{document}

\maketitle

\begin{abstract}

There has been considerable attention devoted to models that learn to jointly infer an expression's syntactic structure and its semantics.
Yet, \citet{NangiaB18} has recently shown that the current best systems fail to learn the correct parsing strategy on mathematical expressions generated from a simple context-free grammar.
In this work, we present a recursive model inspired by \newcite{ChoiYL18} that reaches near perfect accuracy on this task.
Our model is composed of two separated modules for syntax and semantics.
They are cooperatively trained with standard continuous and discrete optimisation schemes.
Our model does not require any linguistic structure for supervision, and its recursive nature allows for out-of-domain generalisation.
Additionally, our approach performs competitively on several natural language tasks, such as Natural Language Inference and Sentiment Analysis.
\end{abstract}

\section{Introduction}
Standard linguistic theories propose that natural language is structured as nested constituents organised in the form of a tree~\citep{Partee:etal:1990}.
However, most popular models, such as the Long Sort-Term Memory network (LSTM)~\citep{hochreiter1997long}, process text without imposing a grammatical structure.
To bridge this gap between theory and practice models that process linguistic expressions in a tree-structured manner have been considered in recent work~\citep{SocherPWCMNP13,TaiSM15,ZhuSG15,BowmanGRGMP16}.
These tree-based models explicitly require access to the syntactic structure for the text, which is not entirely satisfactory.

Indeed, parse tree level supervision requires a significant amount of annotations from expert linguists.
These trees have been annotated with different goals in mind than the tasks we are using them for.
Such discrepancy may result in a deterioration of the performance of models relying on them.
Recently, several attempts were made to learn these models without explicit supervision for the parser~\citep{YogatamaBDGL16, MaillardCY17, ChoiYL18}.
However, \citet{WilliamsDB18} has recently shown that the structures learned by these models cannot be ascribed to discovering meaningful syntactic structure.
These models even fail to learn the simple context-free grammar of nested mathematical operations~\citep{NangiaB18}.

In this work, we present an extension of \citet{ChoiYL18}, that successfully learns these simple grammars while preserving competitive performance on several standard linguistic tasks.
Contrary to previous work, our model makes a clear distinction between the parser and the compositional function.
These two modules are trained with different algorithms, cooperating to build a semantic representation that optimises the objective function.
The parser's goal is to generate a tree structure for the sentence.
The compositional function follows this structure to produce the sentence representation.
Our model contains a continuous component, the compositional function, and a discrete one, the parser.
The whole system is trained end-to-end with a mix of reinforcement learning and gradient descent.
\citet{Drozdov2017} has noticed the difficulty of mixing these two optimisation schemes without one dominating the other.
This typically leads to the ``coadaptation problem'' where the parser simply follows the compositional function and fails to produce meaningful syntactic structures.
In this work, we show that this pitfall can be avoided by synchronising the learning paces of the two optimisation schemes.
This is achieved by combining several recent advances in reinforcement learning.
First, we use input-dependent control variates to reduce the variance of our gradient estimates~\citep{Sheldon_book}.
Then, we apply multiple gradient steps to the parser's policy while controlling for its learning pace using the Proximal Policy Optimization (PPO) of \citet{SchulmanWDRK17}.
The code for our model is publicly available\footnote{\url{https://github.com/facebookresearch/latent-treelstm}}.

\section{Preliminaries}
\label{sec:preliminaries}
In this section, we present existing works on Recursive Neural Networks and their training in the absence of supervision on the syntactic structures.

\subsection{Recursive Neural Networks}
A Recursive Neural Network~(RvNN) has its architecture defined by a directed acyclic graph~(DAG) given alongside with an input sequence~\cite{goller1996learning}.
RvNNs are commonly used in~NLP to generate sentence representation that leverages available syntactic information, such as a constituency or a dependency parse trees~\cite{SocherPHNM11}.

Given an input sequence and its associated~DAG, a~RvNN processes the sequence by applying a transformation to the representations of the tokens lying on the lowest levels of the~DAG.
This transformation, or compositional function, merges these representations into representations for the nodes on the next level of the~DAG.
This process is repeated recursively along the graph structure until the top-level nodes are reached.
In this work, we assume that the compositional function is the same for every node in the graph.

\paragraph{Tree-LSTM.}
\label{eq:tree-lstm}
We focus on a specific type of~RvNNs, the tree-based long short-term memory network~(Tree-LSTM) of~\citet{TaiSM15} and~\citet{ZhuSG15}.
Its compositional function generalizes the LSTM cell of~\citet{hochreiter1997long} to tree-structured topologies, i.e.,
\begin{align*}
\begin{bmatrix} \mathbf{z}\\ \mathbf{i} \\ \mathbf{f}_l\\ \mathbf{f}_r \\ \mathbf{o} \end{bmatrix} &= \begin{bmatrix} \texttt{tanh} \\ \sigma \\ \sigma \\ \sigma \\ \sigma \end{bmatrix} \left( \mathbf{R} \begin{bmatrix}\mathbf{h}_l \\ \mathbf{h}_r\end{bmatrix} + \mathbf{b} \right),\\
\mathbf{c}_p &= \mathbf{z} \odot \mathbf{i} + \mathbf{c}_l \odot \mathbf{f}_l + \mathbf{c}_r \odot \mathbf{f}_r,\\
\mathbf{h}_p &= \texttt{tanh}(\mathbf{c}_p) \odot \mathbf{o},
\end{align*}
where~\(\sigma\) and~\(\texttt{tanh}\) are the sigmoid and hyperbolic tangent functions.
Tree-LSTM cell is differentiable with respect to its recursion matrix~\(\mathbf{R}\), bias~\(\mathbf{b}\) and its input.
The gradients of a Tree-LSTM can thus be computed with backpropagation through structure~(BPTS)~\cite{goller1996learning}.

\subsection{Learning with RvNNs}

A tree-based RvNN is a function~\(f_\theta\) parameterized by a~\(d\) dimensional vector~\(\theta\) that predicts an output~\(y\) given an input~\(x\) and a tree~\(t\).
Given a dataset~\(\mathcal{D}\) of~$N$ triplets~$(x,t,y)$, the parameters of the RvNN are learned with the following minimisation problem:
\begin{align}
  \min_{\theta\in\mathbb{R}^d} \frac{1}{N} \sum_{(x,t,y)\in\mathcal{D}} \ell(f_\theta(x,t),y),
  \label{eq:sup}
\end{align}
where $\ell$ is a logistic regression function.
These models need an externally provided parsing tree for each input sentence during both training and evaluation.
Alternatives, such as the shift-reduce-based SPINN model of~\citet{BowmanGRGMP16}, learn an internal parser from the given trees.
While these solutions do not need external trees during evaluation, they still require tree level annotations for training.
More recent work has focused on learning a latent parser with no direct supervision.

\subsection{Latent tree models}
\label{sec:LatentTreeLearning}

Latent tree models aim at jointly learning the compositional function~$f_\theta$ and a parser without supervision on the syntactic structures~\citep{YogatamaBDGL16, MaillardCY17, ChoiYL18}.
The latent parser is defined as a parametric probability distribution over trees conditioned on the input sequence.
The parameters of this tree distribution~\(p_\phi(.|x)\) are represented by a vector~$\phi$.
Given a dataset~\(\mathcal{D}\) of pairs of input sequences~\(x\) and outputs~\(y\),
the parameters~$\theta$ and~$\phi$ are jointly learned by minimising the following objective function:
\begin{equation}
  \min_{\theta,\phi}\mathcal{L}(\theta,\phi) = {\frac{1}{N}\sum_{(x,y)}
  \ell(\mathbb{E}_{\phi}[f_\theta(x, t)], y)},
  \label{eq:unsup}
\end{equation}
where~$\mathbb{E}_{\phi}$ is the expectation with respect to the~$p_\phi(.|x)$ distribution.
Directly minimising this objective function is often difficult due to expensive marginalisation of the unobserved trees. Hence, when \(\ell\) is a convex function (e.g. cross entropy of an exponential family) usually an upper bound of Eq.~(\ref{eq:unsup}) can be derived by applying Jensen's inequality:
\begin{equation}
\hat{\mathcal{L}}(\theta,\phi) = {\frac{1}{N}\sum_{(x,y)}
  \mathbb{E}_{\phi}[\ell (f_\theta(x, t), y)]}.
\label{eq:bound}  
\end{equation}
Learning a distribution over a set of discrete items involves a discrete optimisation scheme.
For example, the RL-SPINN model of~\citet{YogatamaBDGL16} uses a mix of gradient descent for~$\theta$ and REINFORCE for~$\phi$~\citep{WilliamsDB18}.
\citet{Drozdov2017} has recently observed that this optimisation strategy tends to produce poor parsers, e.g., parsers that only generate left-branching trees.
The effect, called the coadaptation issue, is caused by both bias in the parsing strategy and a difference in convergence paces of continuous and discrete optimisers.
Typically, the parameters~$\theta$ are learned more rapidly than~$\phi$.
This limits the exploration of the search space to parsing strategies similar to those found at the beginning of the training.

\subsubsection{Gumbel Tree-LSTM}
\label{subsection:choi's model}
In their Gumbel Tree-LSTM model,~\citet{ChoiYL18} propose an alternative parsing strategy to avoid the coadaptation issue.
Their parser incrementally merges a pair of consecutive constituents until a single one remains.
This strategy reduces the bias towards certain tree configurations observed with RL-SPINN.

Each word $i$ of the input sequence is represented by an embedding vector.
A leaf transformation maps this vector to pair of vectors~\(\mathbf{r}_i^0{=}(\mathbf{h}_i^0, \mathbf{c}_i^0)\). 
We considered three types of leaf transformations: affine transformation, LSTM and bidirectional LSTM.
The resulting representations form the initial states of the Tree-LSTM.
In the absence of supervision, the tree is built in a bottom-up fashion by recursively merging consecutive constituents $(i, i+1)$ based on merge-candidate scores.
On each level $k$ of the bottom-up derivation, the merge-candidate score of the pair $(i, i+1)$ is computed as follow:
\begin{equation*}
  s_k(i) = \langle \mathbf{q}, \texttt{Tree-LSTM} (\mathbf{r}_i^{k}, \mathbf{r}_{i+1}^{k})\rangle,
\end{equation*}
where~\(\mathbf{q}\) is a trainable query vector and~$\mathbf{r}_i^{k}$ is the constituent representation at position~$i$ after~$k$ mergings.
We merge a pair~$(i^*,i^*+1)$ sampled from the Categorical distribution built on the merge-candidate scores.
The representations of the constituents are then updated as follow:
\begin{equation*}
\mathbf{r}^{k+1}_i =
\begin{cases}
    \mathbf{r}_i^k,           &  i < i^*, \\
    \texttt{Tree-LSTM} (\mathbf{r}_i^{k}, \mathbf{r}_{i+1}^{k})  &  i = i^*, \\
    \mathbf{r}_{i+1}^k        &  i > i^*.
\end{cases}
\end{equation*}
This procedure is repeated until one constituent remains.
Its hidden state is the input sentence representation.
This procedure is non-differentiable. 
\citet{ChoiYL18} use an approximation based on the Gumbel-Softmax distribution~\cite{MaddisonMT16, JangGP16} and the reparametrization trick \cite{KingmaW13}.

This relaxation makes the problem differentiable at the cost of a bias in the gradient estimates~\cite{JangGP16}.
This difference between the real objective function and their approximation could explain why their method cannot recover simple context-free grammars~\cite{NangiaB18}.
We investigate this question by proposing an alternative optimisation scheme that directly aims for the correct objective function.

\section{Our model}
\label{sec:model}
We consider the problem defined in Eq.~(\ref{eq:bound}) to jointly learn a composition function and an internal parser.
Our model is composed of the parser of~\citet{ChoiYL18} and the Tree-LSTM for the composition function.
As suggested in past work~\cite{MnihBMGLHSK16, SchulmanWDRK17}, we added an entropy~$\mathcal{H}$ over the tree distribution to the objective function:
\begin{align}
  \min_{\theta,~\phi}{\hat{\mathcal{L}}}(\theta,\phi)-\lambda\sum_x \mathcal{H}(t~|~x),
  \label{eq:unsup_entropy}
\end{align}
where~$\lambda>0$.
This regulariser improves exploration by preventing early convergence to a suboptimal deterministic parsing strategy.
The new objective function is differentiable with respect to~$\theta$, but not~$\phi$, the parameters of the parser.
Learning~\(\theta\) follows the same procedure with BPTS as if the tree would be externally given.

In the rest of this section, we discuss the optimization of the parser and a cooperative training strategy to reduce the coadaptation issue.

\subsection{Unbiased gradient estimation}
We cast the training of the parser as a reinforcement learning problem.
The parser is an agent whose reward function is the negative of the loss function defined in Eq.~(\ref{eq:bound}).
Its action space is the space of binary trees.
The agent's policy is a probability distribution over binary trees that decomposes as a sequence of~\(K\) merging actions:
\begin{equation}
p_\phi(t|x)=\prod\limits_{k=0}^K \pi_\phi(a_k^i | \mathbf{r}^k),
\label{eq:policy}
\end{equation}
where~\(\mathbf{r}^k=(\mathbf{r}_0^k,\dots,\mathbf{r}_{K-k}^k)\).
The loss function is optimised with respect to~\(\phi\) with REINFORCE~\citep{Williams92}.
REINFORCE requires a considerable number of random samples to obtain a gradient estimate with a reasonable level of variance.
This number is positively correlated with the size of the search space, which is exponentially large in the case of binary trees.
We consider several extensions of REINFORCE to circumvent this problem.

\paragraph{Variance reduction.}
An alternative solution to increasing the number of samples is the control variates method~\citep{Sheldon_book}.
It takes advantage of random variables with known expected values and positive correlation with the quantity whose expectation is tried to be estimated.
Given an input-output pair \((x,y)\) and tree \(t\) sampled from \(p_\phi(t|x)\) , let's define the random variable $G$  as:
\begin{equation}
  G(t) = \ell(f_\theta(x, t), y) \frac{\partial{\log{p_\phi(t|x)}}}{\partial{\phi}}.
\label{eq:reinforce}
\end{equation}
According to REINFORCE, calculating the gradient with respect to $\phi$ for the pair $(x,y)$ is then equivalent to determining the unknown mean of the random variable \(G(t)\)\footnote{Note that while we are computing the gradients using $\ell$, we could also directly optimise the parser with respect to downstream accuracy.}.
Let's assume there is a control variate, i.e., a random variable \(b(t)\) that positively correlates with \(G\) and has known expected value with respect to $p_\phi(.|x)$.
Given \(N\) samples of the \(G(t)\) and the control variate \(b(t)\), the new gradient estimator is:
\begin{equation*}
  G_{\text{CV}} = \mathbb{E}_{p_\phi(t|x)}[b(t)] + \frac{1}{N}\left[\sum_{i=1}^N \left( G(t_i) - b(t_i)\right) \right].
\end{equation*}
A popular control variate, or baseline, used in REINFORCE is the moving average of recent rewards multiplied by the score function~\cite{Sheldon_book}:
\begin{equation*}
b(t) = c \nabla_\phi \log p_\phi(t|x).
\end{equation*}
It has a zero mean under the \(p_\phi(.|x)\) distribution and it positively correlates with $G(t)$.

\paragraph{Surrogate loss.}
REINFORCE often is implemented via a surrogate loss defined as follow:
\begin{equation}
\hat{\mathbb{E}}_t\left[r_\phi(t)\ell(f_\theta(x, t), y)\right],
\end{equation}
where \(\hat{\mathbb{E}}_t\) is the empirical average over a finite batch of samples and \(r_\phi(t)=\frac{p_\phi(t|x)}{p_{\phi_{\text{old}}}(t|x)}\) is the probability ratio with \(\phi_{\text{old}}\) standing for the parameters before the update.

\paragraph{Input-dependent baseline.}
The moving average baseline cannot detect changes in rewards caused by structural differences in the inputs.
In our case, a long arithmetic expression is much harder to parse than a short one, systematically leading to their lower rewards.
This structural differences in the rewards aggravate the credit assignment problem by encouraging REINFORCE to discard actions sampled for longer sequences even though there might be some subsequences of actions that produce correct parsing subtrees.

A solution is to make the baseline input-dependent.
In particular, we use the self-critical training (SCT) baseline of~\citet{RennieMMRG17}, defined as:
\begin{equation*}
  b(t, x)=c_{\theta,\phi}(x)\nabla_\phi \log p_\phi(t~|~x),
\end{equation*}
where \(c_{\theta,\phi}\) is the reward obtained with the policy used at test time, i.e., \(\hat{t}=\argmax p_\phi(t|x)\).
This control variate has a zero mean under the \(p_\phi(t|x)\) distribution and correlates positively with the gradients.
Computing the \(\argmax\) of a policy among all possible binary trees has exponential complexity.
We replace it with a simpler greedy decoding, i.e, a tree $t$ is selected by following a sequence of greedy actions $\hat{a}_k$:
\begin{equation*}
\hat{a}_k = \argmax \pi_\phi(a_k~|~\hat{\mathbf{r}}^k).
\end{equation*}
This approximation is very efficient and computing the baseline requires only one additional forward pass.

\paragraph{Gradient normalization.}
We empirically observe significant fluctuations in the gradient norms.
This creates instability that can not be reduced by additive terms, such as the input-dependent baselines.
A solution is to divide the gradients by a coarse approximation of their norm, e.g., a running estimate of the reward standard deviation~\cite{mnih2014neural}.
This trick ensures that the rewards remain approximately in the unit ball, making the learning process less sensitive to steep changes in the loss.

\subsection{Synchronizing syntax and semantics learning with PPO}
\label{section:sync3.2}
The gradients of the loss function from the Eq.~(\ref{eq:unsup_entropy}) are calculated using two different schemes, BPST for the composition function parameters \(\theta\) and REINFORCE for the parser parameters \(\phi\).
Then, both are updated with SGD.
The estimate of the gradient with respect to \(\phi\) has higher variance compared to the estimate with respect to \(\theta\).
Hence, using the same learning rate schedule does not necessarily correspond to the same real pace of learning.
It is \(\phi\) parameters that are harder to optimise, so to improve training stability and convergence it is reasonable to aim for such updates that does not change the policy too much or too little.
A simple yet effective solution is the Proximal Policy Optimization (PPO) of \citet{SchulmanWDRK17}.
It considers the next surrogate loss:
\begin{equation*}
  \hat{\mathbb{E}}_{t}\left[\max \left\{r_\phi(t)\ell\left(f_\theta(x,t),y\right), r^c_\phi(t)\ell\left(f_\theta(x,t),y\right)\right\}\right],
\end{equation*}
Where \(r^c_\phi(t) = \texttt{clip} \left(r_\phi(t), 1-\epsilon,1+\epsilon\right)\) and \(\epsilon\) is a real number in \((0; 0.5]\).
The first argument of the \(\max\) is the surrogate loss for REINFORCE. 
The clipped ratio in the second argument disincentivises the optimiser  from performing updates resulting in large tree probability changes.
With this, the policy parameters can be optimised with repeated \(K\) steps of SGD to ensure a similar ``pace'' of learning between the parser and the compositional function.

\section{Related work}
Besides the works mentioned in Sec.~\ref{sec:preliminaries} and Sec.~\ref{sec:model}, there is a vast literature on learning latent parsers.
Early connectionist work in inferring context-free grammars proposed stack-augmented models and relied on explicit supervision on the strings that belonged to the target language and those that did not \citep{GilesSCLC89,Sun1990,Das1992,MozerD92}. 
More recently, new stack-augmented models were shown to learn latent grammars from positive evidence alone \citep{JoulinM15}.
In parallel to these, other statistical approaches were proposed to automatically induce grammars from unparsed text~\citep{Sampson:1986,Magerman:Marcus:1990,Carroll:Charniak:1992,Brill:1993,Klein:Manning:2002}.
Our work departs from these approaches in that we aim at learning a latent grammar in the context of performing some given task.

\newcite{SocherPHNM11} uses a surrogate auto-encoder objective to search for a constituency structure, merging nodes greedily based on the reconstruction loss.
\newcite{MaillardCY17} defines a relaxation of a CYK-like chart parser that is trained for a particular task.
A similar idea is introduced in \newcite{LeZ15} where an automatic parser prunes the chart to reduce the overall complexity of the algorithm.
Another strategy, similar in nature, has been recently proposed by \newcite{caio1807}, where Gumbel noise is used with differentiable dynamic programming to generate dependency trees.
In contrast, \newcite{YogatamaBDGL16} learns a Shift-Reduce parser using reinforcement learning.
\newcite{jean1806} further proposes a beam search strategy to overcome learning trivial trees.
On a different vein, \newcite{vlad1809} proposes a quadratic penalty term over the posterior distribution of non-projective dependency trees to enforce sparsity of the relaxation.
Finally, there is a large body of work in Reinforcement Learning that aims at discovering how to combine elementary modules to solve complex tasks \citep{Singh92,michael1807,Sahni17}.
Due to the limited space, we will not discuss them in further details.

\section{Experiments}
We conducted experiments on three different tasks:
evaluating mathematical expressions on the ListOps dataset \citep{NangiaB18},
sentiment analysis on the SST dataset \citep{SocherPWCMNP13}
and natural language inference task on the SNLI \citep{BowmanAPM15} and MultiNLI \citep{Williams:etal:2018} datasets.

\paragraph{Technical details.}
\begin{table*}[t]
\begin{center}
\begin{tabular}{l*{9}{c}}
    \toprule
    \multirow{1}{*}{} & \multicolumn{2}{c}{No baseline} &~~&\multicolumn{2}{c}{Moving average} &~~&\multicolumn{2}{c}{Self critical}  \\
    \cmidrule{2-3}
    \cmidrule{5-6}
    \cmidrule{8-9}
    & No PPO  &  PPO && No PPO  &  PPO && No PPO  &  PPO \\
    \midrule
    \(\min\)   & 61.7 & 61.4 && 61.7 & 59.4 && 63.7 & \bf  98.2  \\
    \(\max\)   & 70.1 & 76.6 && 74.3 & 96.0 && 64.1 & \bf 99.6   \\
    \(\mean\pm\std\)    & 66.2 \(\pm\)3.2 & 66.5 \(\pm\)5.9 && 65.5 \(\pm\) 4.7 & 67.5 \(\pm\)14.3 && 64.0 \(\pm\)0.1 &\bf 99.2 \(\pm\)0.5 \\
    \bottomrule
\end{tabular}
\end{center}
\caption{Accuracy on ListOps test set for our model with three different baselines, with and without PPO. We use $K=15$ for PPO.}
\label{table:listops results}
\end{table*}
\begin{table}[t]
\begin{center}
\begin{tabular}{l cc }
\toprule
Model & Accuracy \\
\midrule
LSTM*  & 71.5\(\pm\)1.5 \\
RL-SPINN*  & 60.7\(\pm\)2.6 \\
Gumbel Tree-LSTM*  & 57.6\(\pm\)2.9 \\
\midrule
Ours  & \bf99.2\(\pm\)0.5\\
\bottomrule
\end{tabular}
\end{center}
  \caption{Accuracy on the ListOps dataset. All models have $128$ dimensions. Results for models with * are taken from \newcite{NangiaB18}.}
\label{table:listops comparison}
\end{table}
For ListOps, we follow the experimental protocol of \newcite{NangiaB18}, i.e., a $128$ dimensional model and a ten-way softmax classifier.
However, we replace their multi-layer perceptron (MLP) by a linear classifier.
The validation set is composed of $1$k examples randomly selected from the training set.
For SST and NLI, we follow the setup of \newcite{ChoiYL18}:
we initialise the word vectors with GloVe300D~\cite{PenningtonSM14} and train an MLP classifier on the sentence representations.
The hyperparameters are selected on the validation set using $5$ random seeds for each configuration.
Our hyperparameters are the learning rate, weight decay, the regularisation parameter $\lambda$, the leaf transformations,
variance reduction hyperparameters and the number of updates \(K\) in PPO.
We use an adadelta optimizer~ \cite{adadelta}.

\subsection{ListOps}
The ListOps dataset probes the syntax learning ability of latent tree models~\citep{NangiaB18}.
It is designed to have a single correct parsing strategy that a model must learn in order to succeed.
It is composed of prefix arithmetic expressions and the goal is to predict the numerical output associated with the evaluation of the expression.
The sequences are made of integers in $[0,9]$ and $4$ operations: \(\texttt{MIN}\), \(\texttt{MAX}\), \(\texttt{MED}\) and \(\texttt{SUM\_MOD}\).
The output is an integer in the range $[0,9]$.
For example, the expression \(\texttt{[MIN 2 [MAX 0 1] [MIN 6 3 ] 5 ]}\) is mapped to the output \(\texttt{1}\).
The ListOps task is thus a sequence classification problem with $10$ classes.
There are $90$k training examples and $10$k test examples.
It is worth mentioning that the underlying semantic of operations and symbols is not provided.
In other words, a model has to infer from examples that $\texttt{[MIN 0 1] = 0}$.

As shown in Table~\ref{table:listops comparison}, the current leading latent tree models are unable to learn the correct parsing strategy on ListOps~\citep{NangiaB18}.
They even achieve performance worse than purely sequential recurrent networks.
On the other hand, our model achieves near perfect accuracy on this task, suggesting that our model is able to discover the correct parsing strategy.
%It is worth noting that we do not use the accuracy as a reward function for the parser, but the loss function of Eq.~\ref{eq:unsup_entropy}.
Our model differs in several ways from the Gumbel Tree-LSTM of~\citet{ChoiYL18} that could explain this gap in performance.
In the rest of this section, we perform an ablation study on our model to understand the importance of each of these differences.

\paragraph{Impact of the baseline and PPO.}
We report the impact of our design choices on the performance in Table~\ref{table:listops results}.
Our model without baseline nor PPO is vanilla REINFORCE.
The baselines only improve performance when PPO is used.
Furthermore, these ablated models without PPO perform on-par with the RL-SPINN model (see Table \ref{table:listops comparison}).
This confirms our expectations for models that fail to synchronise syntax and semantics learning. 

Interestingly, using PPO has a positive impact on both baselines, but accuracy remains low with the moving average baseline.
The reduction of variance induced by the SCT baseline leads to a near-perfect recovery of the good parsing strategy in all five experiments.
This shows the importance of this baseline for the stability of our approach.

\paragraph{Sensitivity to hyperparameters.}
Our model is relatively robust to hyperparameters changes when we use the SCT baseline and PPO.
For example, changing the leaf transformation or dimensionality of the model has a minor impact on performance.
However, we have observed that the choice of the optimiser has a significant impact.
For example, the average performance drops to $73.0\%$ if we replace Adadelta by Adam~\cite{adam}.
Yet, the maximum value out of $5$ runs remains relatively high, $99.0\%$.

\paragraph{Untied parameters.}
As opposed to previous work, the parameters of the parser and the composition function are not tied in our model.
Without this separation between syntax and semantics, it would be impossible to update one module without changing the other.
The gradient direction is then dominated by the low variance signal from the semantic component, making it hard to learn the parser.
We confirmed experimentally that our model with tied parameters fails to find the correct parser and its accuracy drops to~$64.7\%$.

\begin{figure}[t]
\centering
\includegraphics[width=0.5\textwidth]{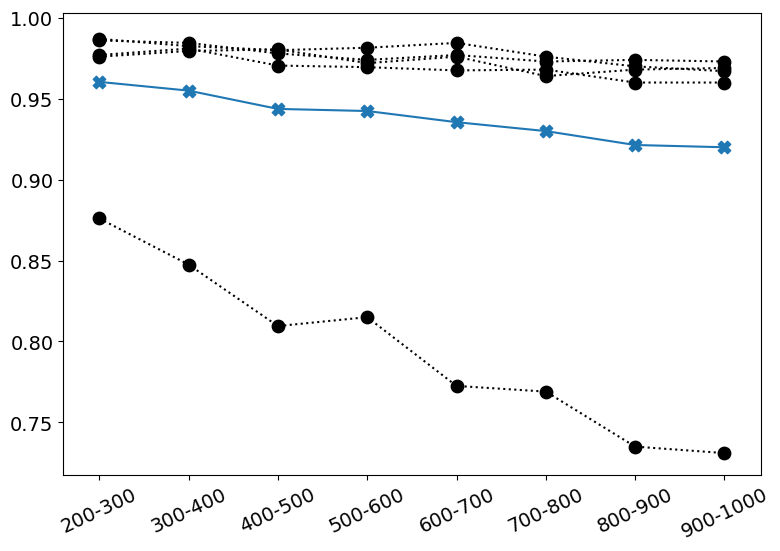}
\vspace*{-7mm}
\caption{Blue crosses depict an average accuracy of five models on the test examples that have lengths within certain range. Black circles illustrate individual models.}
\label{fig:extrapolation}
\end{figure}

\paragraph{Extrapolation and Grammaticality.}
Recursive models have the potential to generalise to any sequence length.
Our model was trained with sequences of length up to \(130\) tokens.
We test the ability of the model to generalise to longer sequences by generating additional expressions of lengths \(200\) to \(1000\).
As shown in Fig.\ref{fig:extrapolation}, our model has a little loss in accuracy as the length increases to ten times the maximum length seen during training.

\begin{figure}[t]
\centering
\includegraphics[width=0.5\textwidth]{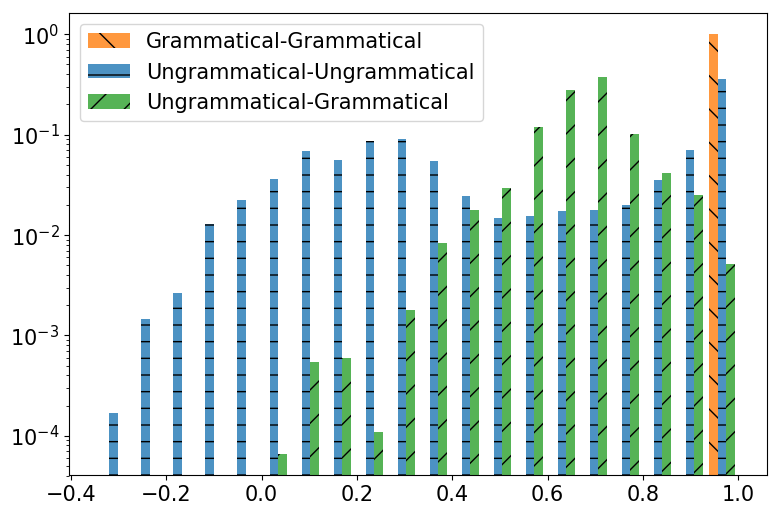}
\vspace*{-3mm}
\caption{The distributions of cosine similarity for elements from the different sets of mathematical expressions. A logarithmic scale is used for y-axis.}
\label{fig:grammaticality}
\end{figure}

On the other hand, we notice that final representations produced by the parser are very similar to each other.
Indeed, the cosine similarity between these vectors for the test set has a mean value of 0.998 with a standard deviation of 0.002.
There are two possible explanations for this observation: either our model assigns similar representations to valid expressions, or it produces a trivial uninformative representation regardless of the expression.
To verify which explanation is correct, we generate ungrammatical expressions by removing either one operation token or one closing bracket symbol for each sequence in the test set.
As shown in Figure \ref{fig:grammaticality}, in contrast to grammatical expressions, ungrammatical ones tend to be very different from each other: ``Happy families are all alike; every unhappy family is unhappy in its own way.''
The only exception, marked by a mode near $1$, come from ungrammatical expressions that represent incomplete expressions because of missing a closing bracket at the end.
This kind of sequences were seen by the parser during training and they indeed have to be represented by the same vector.
These observations show that our model does not produce a trivial representation, but identifies the rules and constraints of the grammar.
Moreover, vectors for grammatical sequences are so different from vectors for ungrammatical ones that you can tell them apart with \(99.99\%\) accuracy by simply measuring their cosine similarity to a randomly chosen grammatical vector from the training set.
Interestingly, we have not observed a similar signal from the vectors generated by the composition function.
Even learning a naive classifier between grammatical and ungrammatical expressions on top of these representations achieves an accuracy of only $75\%$.
This suggests that most of the syntactic information is captured by the parser, not the composition function.

\subsection{Natural Language Inference}
\begin{table}[t]
\setlength{\tabcolsep}{3.5pt} % Default value: 6pt
\begin{center}
\begin{tabular}{l cc }
\toprule
Model & Dim. & Acc. \\
\midrule
\citet{YogatamaBDGL16} & 100 & 80.5\\
\citet{MaillardCY17} & 100& 81.6\\
\citet{ChoiYL18} & 100 & 82.6\\
Ours & 100 & \bf 84.3$\pm$0.3\\
\midrule
\citet{BowmanGRGMP16} & 300 & 83.2 \\
\citet{munkhdalai2017neural} & 300 & 84.6 \\
\citet{ChoiYL18} & 300 &\bf 85.6 \\
\citet{ChoiYL18}\(\dagger\) & 300 & 83.7\\
\citet{ChoiYL18}* & 300 & 84.9\(\pm\) 0.1 \\
Ours & 300 & 85.1$\pm$0.2\\
\midrule
\citet{ChenZLWJI17} & 600 & 85.5\\
\citet{ChoiYL18} & 600 & \bf86.0\\
Ours & 600 & 84.6$\pm$0.2\\
\bottomrule
\end{tabular}
\end{center}
\caption{Results on SNLI. *: publicly available code and hyperparameter optimization was used to obtain results. \(\dagger\): results are taken from \citet{WilliamsDB18}}
\label{table:snli}
\end{table}

\begin{table}[t]
\setlength{\tabcolsep}{3.5pt} % Default value: 6pt
\begin{center}
\begin{tabular}{l cc }
\toprule
Model & Dim. & Acc. \\
\midrule
LSTM\(\dagger\) & 300 & 69.1 \\
SPINN\(\dagger\) & 300 & 67.5 \\
RL-SPINN\(\dagger\) & 300 & 67.4 \\
Gumbel Tree-LSTM\(\dagger\) & 300 & 69.5 \\
\midrule
Ours & 300 & \bf70.7$\pm$0.3\\
\bottomrule
\end{tabular}
\end{center}
\caption{Results on MultiNLI. \(\dagger\): results are taken from \citet{WilliamsDB18}.}
\label{table:mnli}
\end{table}
We next evaluate our model on natural language inference using the Stanford Natural Language Inference (SNLI) \cite{BowmanAPM15} and MultiNLI \cite{Williams:etal:2018} datasets.
Natural language inference consists in predicting the relationship between two sentences which can be either entailment, contradiction, or neutral.
The task can be formulated as a three-way classification problem.
The results are shown in Tables \ref{table:snli} and \ref{table:mnli}.
When training the model on MultiNLI dataset we augment the training data with the SNLI data and use \textit{matched} versions of the development and test sets.
%As in SST experiments we measured efficiency of the parser, it is \(0.54\pm0.3\) on SNLI dataset.
Surprisingly, two out of four models for MultiNLI task collapsed to left-branching parsing strategies. %, while the two other models have an average efficiency of \(0.74\).
This collapse can be explained by the absence of the entropy regularisation and the small number of PPO updates \(K=1\), which were determined to be optimal via hyperparameter optimisation.
As with ListOps, using an Adadelta optimizer significantly improves the training of the model.

\subsection{Sentiment Analysis}

\begin{table}[t]
\setlength{\tabcolsep}{3.5pt} % Default value: 6pt
\begin{center}
\begin{tabular}{l c c}
\toprule
    &  SST-2 & SST-5 \\
\midrule
  \multicolumn{3}{l}{\emph{Sequential sentence representation}}\\
\citet{radford2017learning} & \bf 91.8 & 52.9 \\
\citet{mccann2017learned} & 90.3 & 53.7 \\
\citet{peters2018deep} & - & \bf 54.7\\
\midrule
  \multicolumn{3}{l}{\emph{RvNN based models with external tree}}\\
\citet{SocherPWCMNP13} & 85.4 & 45.7\\
\citet{TaiSM15} & 88.0 & 51.0\\
\citet{munkhdalai2017neural} & 89.3 & 53.1\\
\citet{looks2017deep} & 89.4 & 52.3 \\
\midrule
  \multicolumn{3}{l}{\emph{RvNN based models with latent tree}}\\
\citet{YogatamaBDGL16} & 86.5 & - \\
\citet{ChoiYL18} & 90.7 & 53.7 \\
\citet{ChoiYL18}$^*$ & 90.3\(\pm\)0.5 & 51.6\(\pm\)0.8 \\
\midrule
Ours & 90.2\(\pm\)0.2 & 51.5\(\pm\)0.4 \\
\bottomrule
\end{tabular}
\end{center}
\caption{Accuracy results of models on the SST.
  All the numbers are from~\citet{ChoiYL18} but $^*$ where we used their publicly available code and performed hyperparameter optimization.}
\label{table:sst}
\end{table}

We evaluate our model on a sentiment classification task using the Stanford Sentiment Treebank (SST) of \citet{SocherPWCMNP13}.
All sentences in SST are represented as binary parse trees, and each subtree of a parse tree is annotated with the corresponding sentiment score.
There are two versions of the dataset, with either binary labels, ``negative'' or ``positive'', (SST-2) or five labels, representing fine-grained sentiments (SST-5).
As shown in Table \ref{table:sst}, our results are in line with previous work, confirming the benefits of using latent syntactic parse trees instead of the predefined syntax.

We noticed that all models trained on NLI or sentiment analysis tasks have parsing policies with relatively high entropy.
This indicates that the algorithm does not prefer any specific grammar.
Indeed, generated trees are very similar to balanced ones.
This result is in line with \newcite{shi1808} where they observe that binary balanced tree encoder gets the best results on most classification tasks.

We also compare with state-of-the-art sequence-based models. For the most part, these models are pre-trained on larger datasets and fine-tuned on these tasks.
Nonetheless, they outperform recursive models by a significant margin.
Performance on these datasets is more impacted by pre-training than by learning the syntax.
It would be interesting to see if a similar pre-training would also improve the performance of recursive models with latent tree learning.

\section{Conclusion}
In this paper, we have introduced a novel model for learning latent tree parsers.
Our approach relies on a separation between syntax and semantics.
This allows dedicated optimisation schemes for each module.
In particular, we found that it is important to have an unbiased estimator of the parser gradients and to allow multiple gradient steps with PPO.
When tested on a CFG, our learned parser generalises to sequences of any length and distinguishes grammatical from ungrammatical expressions by forming meaningful representations for well-formed expressions.
For natural language tasks, instead, the model prefers to fall back to trivial strategies, in line with what was previously observed by \newcite{shi1808}.
Additionally, our approach performs competitively on several real natural language tasks.
In the future, we would like to explore further relaxation-based techniques for learning the parser, such as REBAR~\cite{TuckerMMLS17} or ReLAX~\cite{Grathwohl:etal:2017}.
Finally, we plan to look into applying recursive approaches to language modelling as a pre-training step and measure if it has the same impact on downstream tasks as sequential models.

% Include for the final version
\subsubsection*{Acknowledgments}
We would like to thank Alexander Koller, Ivan Titov, Wilker Aziz and anonymous reviewers for their helpful suggestions and comments.

\bibliography{naaclhlt2019}
\bibliographystyle{acl_natbib}

\end{document}